\documentclass[runningheads]{llncs}
\usepackage{graphicx}
\usepackage{comment}
\usepackage{amsmath,amssymb} 
\usepackage{color}
\usepackage{booktabs}
\usepackage{colortbl}
\usepackage{soul}
\usepackage[table]{xcolor}

\usepackage[width=122mm,left=12mm,paperwidth=146mm,height=193mm,top=12mm,paperheight=217mm]{geometry}

\usepackage[unicode=true,
              colorlinks=true,
              linkcolor=blue]{hyperref}

\begin{document}
\pagestyle{headings}
\mainmatter
\def\ECCVSubNumber{1566}  

\title{On the Texture Bias for \\ Few-Shot CNN Segmentation}  

\titlerunning{On the texture bias for few-shot segmentation}
%
\author{Reza Azad\inst{1} \and
Abdur R. Fayjie\inst{1,2} \and
Claude Kauffmann\inst{2} \and
Ismail Ben Ayed\inst{1,2} \and
Marco Pedersoli\inst{1} \and
Jose Dolz\inst{1}}
\authorrunning{R. Azad et al.}
%
\institute{\'ETS Montreal, Canada \and
CRCHUM Montreal, Montreal, Canada\\
\email{rezazad68@gmail.com} \quad
\email{jose.dolz@etsmtl.ca}}
\maketitle

\begin{abstract}


Despite the initial belief that Convolutional Neural Networks (CNNs) are driven by shapes to perform visual recognition tasks, recent evidence suggests that texture bias in CNNs provides higher performing models when learning on large labeled training datasets. This contrasts with the perceptual bias in the human visual cortex, which has a stronger preference towards shape components. Perceptual differences may explain why CNNs achieve human-level performance when large labeled datasets are available, but their performance significantly degrades in low-labeled data scenarios, such as few-shot semantic segmentation. To remove the texture bias in the context of few-shot learning, we propose a novel architecture that integrates a set of Difference of Gaussians (DoG) to attenuate high-frequency local components in the feature space. This produces a set of modified feature maps, whose high-frequency components are diminished at different standard deviation values of the Gaussian distribution in the spatial domain. As this results in multiple feature maps for a single image, we employ a bi-directional convolutional long-short-term-memory to efficiently merge the multi scale-space representations. We perform extensive experiments on three well-known few-shot segmentation benchmarks --Pascal i5, COCO-20i and FSS-1000-- and demonstrate that our method outperforms state-of-the-art approaches in two datasets under the same conditions.

\end{abstract}

\section{Introduction}

Deep models, and particularly convolutional neural networks (CNNs), have shown an impressive performance in many visual recognition tasks, including semantic segmentation \cite{long2015fully}. However, their extreme hunger for labeled training data strongly limits their scalability to novel classes and reduces their applicability to rare categories. Few-shot learning \cite{ravi2016optimization,finn2017model} has appeared as an appealing alternative to train deep models in a low-labeled data scenario. 
In this setting, the model is trained to accommodate for novel categories with only a handful of labeled images, typically known as \textit{support} images. In few-shot segmentation approaches, the learned knowledge from the \textit{support} images is typically fed into a parametric module to guide the segmentation of the unseen images, referred to as \textit{queries}.

Recent works have demonstrated that the CNN bias towards recognizing textures rather than shapes introduces several benefits under the standard learning paradigm \cite{geirhos2018imagenet,brendel2019approximating}, which contrasts with the inductive bias found in the human visual cortex, that is driven by shapes \cite{landau1988importance}. This does not represent a problem when training and testing classes are drawn from the same distribution in large-labeled datasets. Nevertheless, in low-labeled data regime, the difference on perceptual biases poses difficulties to CNNs to mimic human performance, particularly if there exists a distributional shift between training and testing classes, such as in the few-shot learning scenario \cite{ringer2019texture}. 

 

\begin{figure*}[]
\centering
\includegraphics[width=.99 \textwidth]{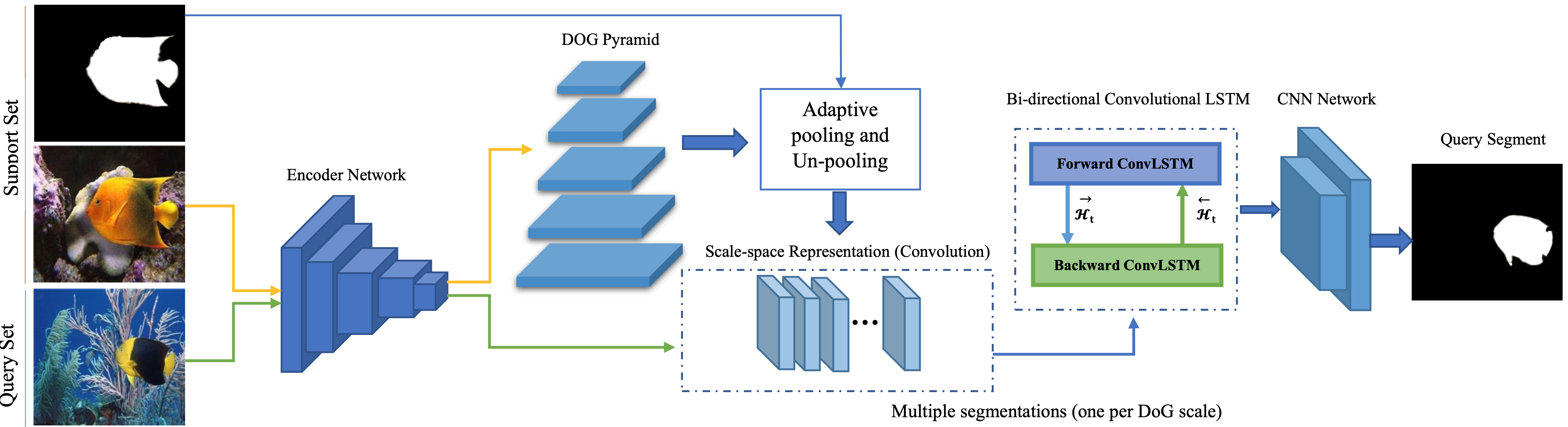}
\caption{Overview of the proposed method (DoG-LSTM) for few-shot segmentation. It first applies a pyramid of difference of gaussians (DoG) on the learned \textit{support} features to attenuate high-frequency local components on the feature space. To perform segmentation on a \textit{query} image, the multiple scale-space \textit{support} representations are combined with the \textit{query} features, and later fed as input to a bi-directional convLSTM. The convLSTM merges the information from multiple representations and generates the final \textit{query} segmentation map.} \label{fig:BConvLSTM}
\vspace*{-\baselineskip}
\label{fig:structure}
\end{figure*}

Thus, we argue that attenuating high-frequency local components in the feature space yields to a better generalization 
to unseen classes in the context of few-shot semantic segmentation. Our motivation is inspired by the findings in \cite{geirhos2018imagenet}, who showed that CNNs have a strong texture inductive bias that limits their ability to leverage useful low-frequency (e.g, shape) information. Although they show that the representational power of CNN can be improved if CNNs are forced to use shape information (by modifying input images), how to design efficient algorithms that allow CNNs to meaningfully use low-frequency information remains an open problem. We tackle this issue by proposing a novel architecture (Fig. \ref{fig:structure}) which integrates a set of difference of gaussians (DOGs) \cite{lowe2004distinctive} on the feature representations. At each scale-space of the DOGs, the original high-frequency signals are attenuated differently, according to the standard deviation values, $\sigma$, employed to model the Gaussian distribution in the spatial domain, which results in multiple versions of the feature maps for a single image. The DoG at two near standard deviations ($\sigma_1$ and $\sigma_2$) will smooth out the features, reducing textural information that the feature extractor may have propagated. Then, following the literature on few-shot segmentation, we generate class representative prototypes from the learned representations, with the difference that in our setting we have multiple prototypes per image, i.e., one at each scale-space of the DOG. Thus, for each query image, our model produces an ensemble of segmentations, each one associated with a prototype. To generate the final prediction, we cast the problem into a sequential segmentation task, where each segmentation on the ensemble represents a time-point. To efficiently fuse temporal, i.e., multiple segmentation masks, and spatial features we resort to a Bi-directional convolutional long-short-term memory (BConvLSTM) \cite{Song_2018_ECCV}, which bidirectionally encourages information exchange between LSTM units. Furthermore, in the k-shot setting, our approach learns a parametric fusion of the different support images by jointly analyzing their contribution.

Our contributions can be summarized as follows: (1) we propose to reduce the texture bias in CNNs in the few-shot segmentation task by attenuating high-frequency local components on the feature space, (2) to merge the multiple segmentations produced at different scale-space representations we reformulate the problem as a sequential segmentation task and employ a bi-directional convLSTM to efficiently fuse all the information, and (3) we report very competitive results on few-shot segmentation across several public benchmarks, outperforming most recent literature while keeping a light architecture. 

\section{Related Work}


\paragraph{\textbf{Few-shot segmentation.}} Pioneer works on few-shot semantic segmentation \cite{shaban2017one,dong2018few,rakelly2018conditional} incorporated two independent branches: a conditional branch that generates the prototypes (e.g., embedding) from the \textit{support} set, and a segmentation branch, which takes the learned prototypes and the \textit{query} image as input and produces the segmentation masks. More recently, researchers have unified these dual-branch architectures into a single branch network which can derive better guidance features with the addition of a masked average pooling layer \cite{zhang2018sg,siam2019amp,wang2019panet,nguyen2019feature}. For example, a similarity guidance module is integrated in \cite{zhang2018sg} to recalibrate the query feature map based on a similarity score between the representative prototype and each spatial location on the query features. 
In \cite{siam2019amp}, authors present an approach to generate the weights of the final segmentation layer for the novel classes via imprinting. 
Other works interchange support and query images for prototype alignment regularization \cite{wang2019panet} or to concurrently make predictions \cite{liu2020crnet} with the goal of achieving better generalization.
Nguyen et al. \cite{nguyen2019feature} integrated a regularization that estimates feature relevance by encouraging jointly high-feature activations on the foreground and low-feature activations on the background. Deep attention has also been exploited to learn attention weights between support and query images for further label propagation \cite{hu2019attention,zhang2019canet}. More recently, some researchers have adopted graph CNNs to establish more robust correspondences between support and query images and enrich the prototype representation \cite{liu2020part,wangfew}. Our work differs from previous approaches from a motivation and methodological perspective. While most of the current literature focuses on learning better prototypical representations or iteratively refining these, we approach this problem under the perspective of reducing the inductive texture bias of CNNs. Thus, from a methodological point of view, our approach is the first attempt to integrate a pyramidal set of DoG to address the problem of texture bias.




\paragraph{\textbf{Semantic segmentation with conv LSTM.}}  

Conv Long-Short Term Memory (LSTM) was presented in \cite{xingjian2015convolutional} to address the limitations of LSTMs in tasks such as semantic segmentation, where the learned intermediate representations of the input images must preserve the spatial information. Particularly, convLSTM addresses this by integrating a convolution operator in the state-to-state and input-to-state transitions. 
In the context of image segmentation on 
3-dimensional data, e.g. videos or medical imaging, convLSTMs are integrated to encode the spatial-temporal relationships between frames or slices \cite{chen2016combining,valipour2017recurrent,Song_2018_ECCV,zhang2018multi}. If only 2D images are available instead, an alternative is to leverage convLSTM for multi-level feature fusion \cite{li2018referring,azad2019bi}. Li et al. \cite{li2018referring} employed convLSTM units to progressively refine the segmentation masks from high-level to low-level features. In \cite{azad2019bi}, features derived from the skip connections in the encoding path of UNet \cite{ronneberger2015u} were non-linearly fused with their corresponding features in the decoding path by employing a bi-directional convLSTM, instead of a simple concatenation. In a related work, Hu et al. \cite{hu2019attention} employ a ConvLSTM to merge multiple segmentations in a $k$-shot scenario ($k>1$), where each segmentation is generated from a different support image. This differs from our work, where our goal is to fuse the segmentations from a single support image ($k=1$) derived from multiple scale-space representations. Furthermore, we use a bidirectional ConvLSTM to foster the exchange of information between the forward and backward path of each recurrent module.

\section{Methodology}

\subsection{Problem Formulation}

Following the standard notation and set-up in few-shot semantic segmentation, we define three datasets: a training set $D_{train}=\{(X_i^t,Y_i^t)\}_{i=1}^{N_{train}}$, a support set $D_{support}=\{(X_i^s,Y_i^s)\}_{i=1}^{N_{support}}$, and a test set $D_{test}=\{(X_i^q)\}_{i=1}^{N_{test}}$. In this setting, $X_i \in \mathbb{R}^{H\times W \times 3}$ denotes an RGB image, with $H$ and $W$ being the height and the width of the image, respectively, and $Y_i \in \{0,1\}^{H\times W}$ is its corresponding pixel-level mask. Furthermore, each set contains $N$ images. The classes, denoted as $c \in C$, are shared among the support and test set, and are disjoint with the training set, i.e.,  $\{C_{train}\} \cap \{C_{support}\}=\emptyset$. 

The purpose of few-shot learning is to train a neural network $f_{\theta}(\cdot)$ on the training set $D_{train}$ to have the ability to segment a novel class $c \notin C_{train}$ on the test set $D_{test}$, based on $k$ references from the support set $D_{support}$. To reproduce this mechanism during training, the network is trained on $D_{train}$ following the episodic paradigm \cite{vinyals2016matching}. 
Specifically, assuming a $c$-way $k$-shot learning task, each episode is generated by sampling: (1) a support training set $D_{train}^{\mathcal{S}}=\{(X_s^t,Y_s^t(c))\}_{s=1}^{k} \subset D_{train}$ for each class $c$, where $Y_s^t(c)$ is the binary mask for the class $c$ corresponding to the image $X_s^t$ and (2) a query set $D_{train}^{\mathcal{Q}}=\{X_q^t,Y_q^t(c)\} \subset D_{train}$, where $X_q^t$ is the query image and $Y_q^t(c)$ its corresponding binary mask for the class $c$. The input of the model is composed of the support training set and the query image, $f_{\theta}(D_{train}^{\mathcal{S}},X_q^t)$, which are employed to estimate the segmentation mask for the class $c$ in the query image, $\hat Y_q^t(c)$. Then, the neural network parameters $\theta$ are optimized by employing an objective function between $Y_q^t(c)$ and $\hat Y_q^t(c)$\footnote{Typically the standard cross-entropy loss function is employed in the few-shot segmentation literature.}. During the testing phase, the model $f_\theta(\cdot)$ is evaluated on the test set $D_{test}$ given $k$ images from the support set $D_{support}$.

\subsection{Removing Texture Bias}


Recent findings suggest that perceptual bias on CNNs do not correlate with those in the human visual cortex \cite{geirhos2018imagenet}, which may limit the performance of these models in low-labeled data scenarios \cite{ringer2019texture}. Inspired by this, we propose to reduce the texture bias of CNNs in the context of few-shot segmentation. To achieve this, we integrate a set of difference of gaussians (DoGs) \cite{lowe2004distinctive} into the learned feature space to attenuate high-frequency local components, i.e., texture. First, we use a CNN to encode the input images into the latent space, resulting in $F_s \in \mathbb{R}^{W' \times H' \times M}$ and $F_q \in \mathbb{R}^{W' \times H' \times M}$ for the support and query samples. The variables $W'$, $H'$ and $M$ represent the width, height and feature dimensionality on the latent space, respectively. To encode the high-frequency information during training, we apply a DoG on each channel $m \in M$ of the feature map from the support samples $F_s$, which can be formulated as:
\begin{equation}
  G_{s} = \Gamma_{\sigma_1,\sigma_2} (F_s)  = (F_s^m \ast\frac{1}{2 \pi \sigma_2^2}\exp^{-\frac{x^2+y^2}{2 \sigma_2}}) - 
  (F_s^m \ast\frac{1}{2 \pi \sigma_1^2}\exp^{-\frac{x^2+y^2}{2 \sigma_1}}), \forall m \in M
  \label{eq:DOG_single}
\end{equation}

where $\sigma_1$ and $\sigma_2$ are ($\sigma_2 > \sigma_1$) are the variance of the Gaussian filters, $x$ and $y$ represent the spatial position in the encoded feature space and $\ast$ denotes the convolution operator. To encode different frequency information we apply a pyramid of DoGs with increasing $\sigma$ values, similar to \cite{lowe2004distinctive}. This results in $L$ level representations ($L=4$) for each support sample (See Fig. \ref{fig:BConvLSTM}), where the novel feature maps at each level ($l \in L$) can be denoted as $G_{s}^l \in \mathbb{R}^{W' \times H' \times M}$.

\begin{figure*}[]
\centering
\includegraphics[width=0.91 \textwidth]{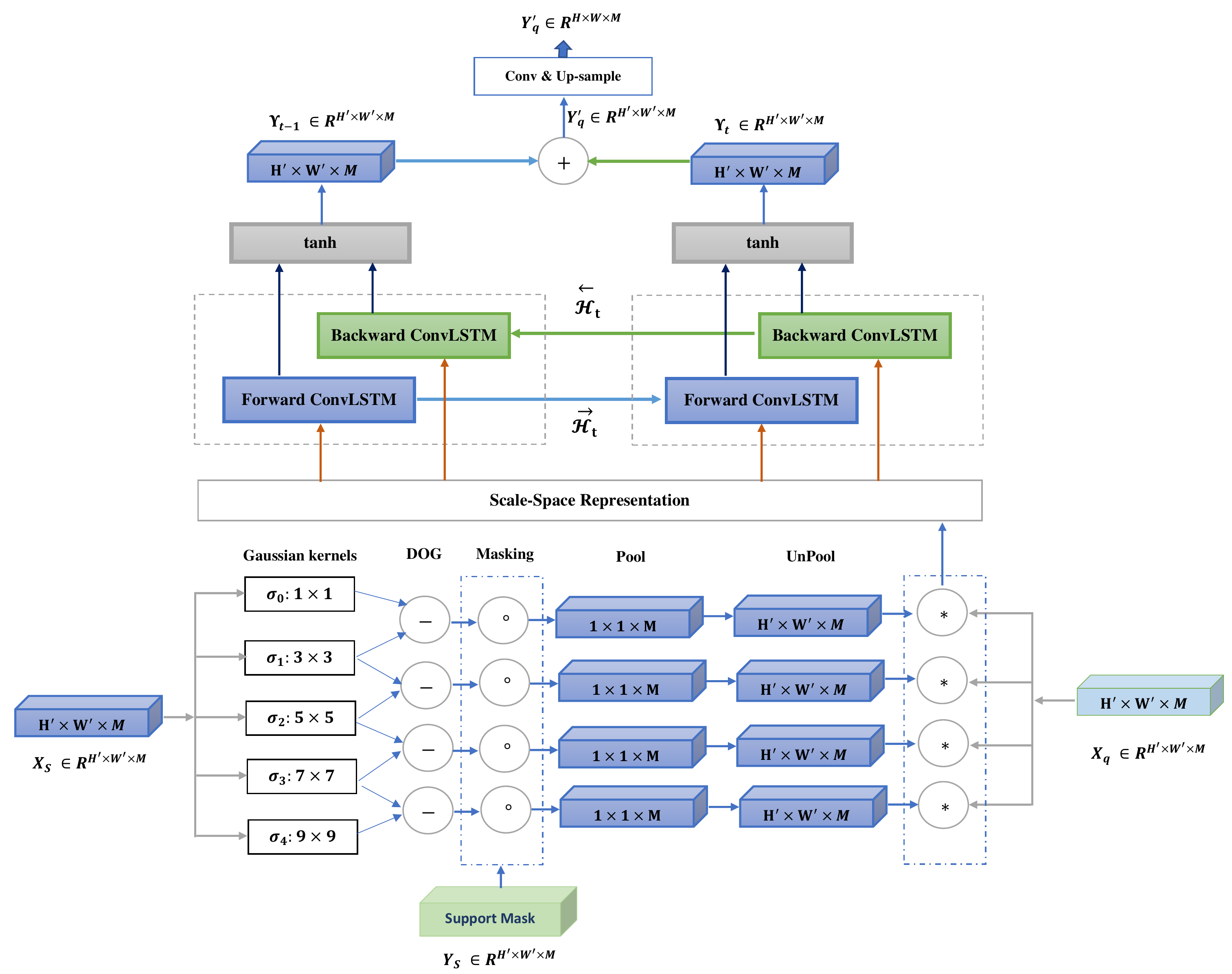}
\caption{The scale-space encoding block in the proposed method. } \label{fig:BConvLSTM}
\end{figure*}

Support images can contain cluttered background, as well as multiple object categories. Thus, we need to find a representative embedding $f_s$ that corresponds exclusively to the target class. 
Since we have $L$ feature representations, each of them encoding different high-frequency local components, we generate $L$ prototypes per class. To obtain the class prototypes, the novel encoded feature maps at each scale $G_s^l$ are averaged over the known foreground regions in the support mask $Y_s(c)$. Thus, at each level we can estimate $f_s^l$ as:
\begin{align}
  f_s^l = \frac{1}{| \tilde{Y}_s(c)|} \sum_{i=1}^{W'H'}G^l_s\tilde{Y}_s(c)
  \label{eq:proto}
\end{align}
where the support mask $Y_s(c)$ is down-sampled to $\tilde{Y}_s(c) \in \{ 0,1\}^{H' \times W'}$ to match the spatial resolution of the feature maps $G^l_s$ and $|\tilde{Y}_s(c)|=\sum_i\tilde{Y}_{s,i}(c)$ is the number of foreground locations in $\tilde{Y}_s(c)$. Then, each prototype is unpooled to the same spatial resolution as the query features $F_q$ and the upsampled prototypes are convolved with $F_q$. We then define the scale-space representation (SSR), which will serve as input signal of the BConvLSTM. This representation can be formulated as a convolution operation between the class representative feature maps at each scale-space and the feature maps derived from the query image: 
\begin{align}
 SSR = \{BN(\psi_{s}^{l} \ast F_q)\}, \forall l \in L
  \label{eq:SSR}
\end{align}
where $\psi_{s}^{l}$ are the upsampled prototypes $f^l_s$, and $BN$ denotes a batch normalization layer.






\subsection{Encoding Scale-Space Representation}

Fusion of the query features $F_q$ with the multi-scale class representations from the support features $\psi_s^l$ produces $L$ joint feature maps, one at each scale-space representation. While logical or average operations may be a straightforward solution to obtain a unique representation, they fail to exploit the inner relationship between sequential scale-space representations. To efficiently solve this, we reformulate the problem as a sequential task, and integrate a bidirectional convolutional long short term memory (BConvLSTM) \cite{Song_2018_ECCV} on the output of the CNN architecture (Fig. \ref{fig:BConvLSTM}). Even though LSTM have been proposed to deal with sequential problems, this sequential processing strategy may fail to explicitly encode the spatial correlation, since they use full connections in input-to-state and state-to-state transitions.
To overcome this limitation, ConvLSTM was proposed in \cite{xingjian2015convolutional}, which leverages convolution operations into input-to-state and state-to-state transitions instead. Specifically, three gating functions are calculated in the ConvLSTM, which are defined as:

\begin{align}
  i_t = \sigma(\textbf{W}_{xi} \ast \mathcal{X}_t + \textbf{W}_{hi} \ast \textbf{H}_{t-1} + \textbf{W}_{ci} \circ \textbf{C}_{t-1} + b_i)
  \label{eq:lstm1}
\end{align}
\begin{align}
  f_t = \sigma(\textbf{W}_{xf} \ast \mathcal{X}_t + \textbf{W}_{hf} \ast \textbf{H}_{t-1} + \textbf{W}_{cf} \circ \textbf{C}_{t-1} + b_f)
  \label{eq:lstm2}
\end{align}
\begin{align}
  o_t = \sigma(\textbf{W}_{xo} \ast \mathcal{X}_t + \textbf{W}_{ho} \ast \textbf{H}_{t-1} + \textbf{W}_{co} \circ \textbf{C}_{t-1} + b_{0} )
  \label{eq:lstm3}
\end{align}
where $\mathcal{X}_t$ and $\textbf{H}_t$ denote the input (i.e., $SSR$ in eq. (\ref{eq:SSR})) and hidden state at time $t$, respectively, and $b$ is used to represent the bias term in each state. Similarly, $\textbf{W}_{x}$, $\textbf{W}_{h}$ and $\textbf{W}_{c}$ represent the set of learnable parameters. 
Last, `$\circ$' denotes the Hadamard product.
The LSTM module generates a new proposal for the cell state by looking at the previous $\textbf{H}$ and current $\mathcal{X}$, resulting in:
\begin{align}
  \tilde{\textbf{C}}_t = \tanh(\textbf{W}_{xc} \ast \mathcal{X}_t + \textbf{W}_{hc} \ast \textbf{H}_{t-1} + b_c)
  \label{eq:lstm4}
\end{align}
Now we linearly combine the newly generated proposal $\tilde{\textbf{C}}_t$ with the previous state $\textbf{C}_{t-1}$ to generate the final cell state in the recurrent model:
\begin{align}
  \textbf{C}_t = f_t \circ \textbf{C}_{t-1} + i_t \circ \tilde{\textbf{C}}_t
  \label{eq:lstm5}
\end{align}
Finally, the new hidden state $\textbf{H}$ can be estimated as:
\begin{align}
  \textbf{H}_t = o_t \circ \tanh(\textbf{C}_t)
  \label{eq:lstm6}
\end{align}
Inspired by \cite{Song_2018_ECCV}, we employ in this work a BConvLSTM to encode the different scale-space representations (SSR) at the output of the convolutional network (Fig. \ref{fig:BConvLSTM}). The bidirectional modules with forward and backward paths allow to strength the spatio-temporal information exchanges between the two sides, facilitating the memorization of both past and future sequences. This contrasts with the standard convLSTM, where only the dependencies on the forward direction are employed for the predictions. Thus, the output prediction for a query image $X^{q}$ is given at the output of the BConvLSTM, which is defined as:
\begin{align}
  \hat{Y}^q = \tanh(\textbf{W}_y^{\overrightarrow{\textbf{H}}} \ast \overrightarrow{\textbf{H}} + \textbf{W}_y^{\overleftarrow{\textbf{H}}} \ast \overleftarrow{\textbf{H}} + b) 
  \label{eq:lstm6}
\end{align}
where $\overrightarrow{\textbf{H}}$ and $\overleftarrow{\textbf{H}}$ represent the hidden states of the forward and backward convLSTM units, respectively, and $b$ is the bias term. 
Last, the output of the BConvLSTM is passed through a series of convolutions, followed by upsampling and batch normalization layers to produce the final segmentation masks in the original input image resolution.




\subsection{k-shot Segmentation}
To fuse the information from several support images ($k>1$), most previous works estimate the class prototype $\psi$ by simply taking the average of the representation vectors among $k$ samples (non-parametric approach) \cite{shaban2017one,zhang2018sg,siam2019amp}. Nevertheless, this strategy assumes that each $k$ sample has equal importance, and thus fails to provide a robust category representation when dealing with noisy or corrupted samples. To deal with this limitation, we propose to use a non-linear parametric method to further improve the model performance on the $k$-shot setting. The key idea is to first generate the embedded representation between the query and each $k$ support samples and then apply BConvLSTM on these representations to get the final representation in a non-linear parametric fashion. Moving $k$-shot setting inside the scale-space representation gives the BConvLSTM more freedom to generate better representations using various samples. 

\subsection{Weakly-supervised Few-shot Segmentation}

To push further the idea of training with very few supervision, we explore the performance of our method when other forms different than full-supervision , i.e., full pixel-level masks, 
are available. Particularly, we investigate bounding box annotations, 
which are less time-consuming to obtain than exhaustive segmentation masks. In this context, we relax the support mask by considering all the area inside the bounding box as the foreground. We show in the experiments 
that, compared to pixel-level annotations, our model achieves very competitive results by employing sparse support annotations.

\section{Experiments}
In this section, we present the datasets employed to evaluate our method and the experimental setting in our experiments. We then report the results compared to state-of-the-art few-shot segmentation approaches, demonstrating the benefits of our method.

\subsection{Datasets}

We perform extensive evaluations on three few-shot semantic segmentation benchmarks, i.e., PASCAL-5$^i$, FSS-1000 and COCO, following standard procedures in the literature. Details are given in \textit{Supplemental materials}.

\subsection{Experimental Set-up}

\subsubsection{Network and implementation details.} 
We employ VGG \cite{simonyan2014very} and ResNet-101 pre-trained on ImageNet as the backbones for feature extractor. The proposed model is trained end-to-end by using Adam \cite{kingma2014adam} for 50K episodes with a batch size of 5. The initial learning rate is set to 10$^{-4}$ and reduced by 10$^{-1}$ at every 10K iterations.  
The work is carried out using one NVidia Titan X GPU. The code is written in Keras with tensorflow as backend and the code is publicly available at \url{https://github.com/rezazad68/fewshot-segmentation}



\subsubsection{Evaluation protocol.} 

To evaluate the performance of the few-shot segmentation models, we employ the average IoU over all classes (mIoU). As pointed out in \cite{zhang2019canet}, the mIoU is a better metric, compared to background-foreground IoU (FB-IoU), in the context of few-shot semantic segmentation for several reasons. First, if a given image contains very small objects, the model may completely fail to segment those objects. Nevertheless, the background IoU can still be very high, which misleads information about the real performance of the model. And second, FB-IoU is more suitable for binary segmentation problems, such as video or foreground segmentation, while our purpose is on semantic segmentation.



\subsection{Results}
\definecolor{gray}{rgb}{0.93,0.93,0.93}

\subsubsection{Comparison with state-of-the-art.}
Comparison of the proposed model with state-of-the-art methods on the FSS-1000 and PASCAL-5$^i$ datasets is reported in Tables \ref{table:1000Class} and \ref{table:pascal5}, respectively\footnote{Results on COCO are given in Supplemental Material.}. Results in Table \ref{table:1000Class} show that the proposed model outperforms the state-of-the-art methods in both 1-shot and 5-shot settings employing the same backbone, i.e., VGG. Particularly, in the 1-shot task, our method achieves a significant improvement of 5.5\% over the second best performing model. In the case of 5-shot learning, we found that fusing the segmentations from the different supports in a non-parametric way brings nearly 1\% of improvement with respect to the 1-shot setting. Nevertheless, combining the 5 support segmentations in a parametric fashion, i.e., with BConvLSTM, increases the mIoU by 2.5\%. It is noteworthy to mention that the method in DAN \cite{wangfew} uses ResNet-101 as backbone, which might explain the differences between the different methods.

\begin{table}[h!]
\scriptsize
\begin{center}
\caption{Results of 1-way 1-shot and 1-way 5-shot segmentation on the FSS-1000 data set employing the mean Intersection Over Union (mIoU) metric. Best results in bold.}
\label{table:headings}
\begin{tabular}{llcc}
\hline\noalign{\smallskip}
 Method & mIoU\\
\hline
& \multicolumn{1}{c}{1-shot}  \\
\hline
\noalign{\smallskip}
OSLSM \cite{shaban2017one} & 70.3 \\
co-FCN \cite{rakelly2018conditional} & 71.9 \\
FSS-1000 \cite{wei2019fss} & 73.5 \\
FOMAML \cite{hendryx2019meta} & 75.2 \\
Baseline  & 74.2 \\
Baseline+DoG & 78.7 \\
Baseline+DoG+BConvLSTM  & \bf 80.8 \\
\hline
DAN \cite{wangfew} (ResNet-101) & 85.2 \\
\hline
& \multicolumn{1}{c}{5-shot}  \\
\hline\noalign{\smallskip}
OSLSM \cite{shaban2017one} & 73.0\\
co-FCN \cite{rakelly2018conditional} & 74.3\\
FSS-1000 \cite{wei2019fss} & 80.1\\
FOMAML+regularization \cite{hendryx2019meta} &80.6 \\
FOMAML+regularization+UHO \cite{hendryx2019meta} & 82.2 \\
Baseline+DoG+BConvLSTM (non-param) & 81.7 \\
Baseline+DoG+BConvLSTM (param) & \bf 83.4 \\
\hline
DAN \cite{wangfew} (ResNet-101) & 88.1 \\
\hline

\end{tabular}
\label{table:1000Class}
\end{center}
\vspace{-6mm}
\end{table}

\begin{table*}[h!]
\vspace{-0mm}
\scriptsize
\centering
\begin{center}
\caption{Results of 1-way 1-shot and 1-way 5-shot segmentation on PASCAL-5$^i$ data set employing the mean Intersection-Over-Union (mIoU) metric. Best results for each backbone architecture are highlighted in bold. We employ $\nabla$ to denote the difference between 1- and 5-shot settings.}
\label{table:headings}
\begin{tabular}{l|cccccc|ccccc|c}
\toprule
\noalign{\smallskip}
& \multicolumn{4}{c}{1-shot} & &\multicolumn{4}{c}{5-shot} & \\
\midrule
Method & & fold$^1$ & fold$^2$ & fold$^3$ & fold$^4$ & Mean & fold$^1$ & fold$^2$ & fold$^3$ & fold$^4$ & Mean & $\nabla$\\
\noalign{\smallskip}
\midrule
   & \multicolumn{8}{c}{Backbone (VGG 16)}       \\
 \midrule
\noalign{\smallskip}

OSLSM \cite{shaban2017one} & BMVC'18 &33.6 & 55.3 & 40.9 & 33.5 & 40.8 & 35.9 & 58.1 & 42.7 & 39.1& 43.9 & 3.1\\
co-FCN \cite{rakelly2018conditional} & ICLRW'18 &36.7 & 50.6 & 44.9 & 32.4 & 41.1 & 37.5 & 50.0 & 44.1 & 33.9 & 41.4 & 0.3 \\

AMP \cite{siam2019amp} & ICCV'19 &41.9 & 50.2 & 46.7 & 34.7 & 43.4 & 41.8 & 55.5 & 50.3 & 39.9 & 46.9 & 3.5  \\
PANet \cite{wang2019panet} & ICCV'19 &42.3 & 58.0 & 51.1 & 41.2 & 48.1 & 51.8 & 64.6 & \textbf{59.8} & 46.5 & 55.7 & 7.6 \\
FWB\cite{nguyen2019feature} & ICCV'19 &47.0 & 59.6 & 52.6 & 48.3 & 51.9 & 50.9 & 62.9 & 56.5 & 50.1 & 55.1 & 3.2 \\
Meta-Seg \cite{cao2019meta} & IEEE Access'19&42.2 & 59.6 & 48.1 & 44.4 & 48.6 & 43.1 & 62.5 & 49.9 & 45.3 & 50.2 & 1.6  \\
MDL \cite{dong2019multi} & COMPSAC'19 &39.7 & 58.3 & 46.7 & 36.3 & 45.3 & 40.6 & 58.5 & 47.7 & 36.6 & 45.9 & 0.6  \\
SG-One \cite{zhang2018sg}& IEEE SMC'20 &40.2 & 58.4 & 48.4 & 38.4 & 46.3 & 41.9 & 58.6 & 48.6 & 39.4 & 47.1 & 0.8 \\
OS$_{Adv}$ \cite{yang2020recognizing} & Inf. Sci.'20 &46.9 & 59.2 & 49.3 & 43.4 & 49.7 & 47.2 & 58.8 & 48.8 & 47.4 & 50.6 & 0.9  \\
ARNet \cite{li2020arnet} & ICASSP'20 & 42.6 & 59.5 & 50.2 & 40.2 & 48.1 & 43.3 & 59.8 & 51.7 & 41.4 & 49.1 & 1.0 \\

CRNet \cite{liu2020crnet} & CVPR'20 & - & - & - & - & 55.2 & - & - & - & - & 58.5 & 3.3  \\
FSS-1000 \cite{wei2019fss} & CVPR'20 &- & - & - & - & - & 37.4 & 60.9 & 46.6 & 42.2 & 56.8 & --  \\
RPMM \cite{liu2020part} & ECCV'20 &47.1 & 65.8 & 50.6 & 48.5 & 53.0 & 50.0&  66.5 & 51.9 & 47.6 & 54.0 & 1.0\\
PFNet \cite{tian2020prior} & TPAMI'20 & \bf 56.9 & \bf 68.2 & 54.4 & 52.4 & \bf 58.0 & \bf 59.0 & 69.1 & 54.8 & 52.9 & 59.0 & 1.0 \\
\bf Proposed & - & 56.2 & 66.0 & \textbf{56.1} & \textbf{53.8} & \textbf{58.0} & 57.5 & \textbf{70.6} & 56.6 & \textbf{57.7} & \textbf{60.6} & 2.6\\
 \midrule
& \multicolumn{8}{c}{Backbone (ResNet-50)}       \\
 \midrule
 CANet  \cite{zhang2019canet} & CVPR'19 &52.5 & 65.9 & 51.3 & 51.9 &  55.4 & 55.5 & 67.8 & 51.9 & 53.2  & 57.1 & 1.7 \\
PGNet \cite{zhang2019pyramid}  & ICCV'19 & \bf 56.0 & 66.9 & 50.6 & 50.4 &  56.0 & 57.7 & 68.7 & 52.9 & 54.6  & 58.5 & 2.5 \\
LTM \cite{yang2020new} & MMM'20 &52.8 & \bf 69.6 & 53.2 &  52.3 &  57.0 &  57.9 & \bf 69.9 & 56.9 & \bf 57.5 & 60.6 & 3.6\\
CRNet \cite{liu2020crnet} & CVPR'20 & - & - & - & - & 55.7 & - & - & - & - & 58.8 & 2.9  \\
PPNet \cite{liu2020part}* & ECCV'20 &47.8 & 58.8 & 53.8 & 45.6 & 51.5 & \bf 58.4 & 67.8 & \bf 64.9 &56.7 & \bf 62.0 & 10.5  \\
RPMM \cite{yang2020prototype}& ECCV'20 &55.2 & 66.9 & 52.6 & 50.7 & 56.3 & 56.3 & 67.3 &  54.5 & 51.0 & 57.3 & 1.0 \\
PFNet \cite{tian2020prior} & TPAMI'20 & \bf 61.7 &  69.5 & \bf 55.4 & \bf 56.3 & \bf 60.8 & 63.1 & 70.7 & 55.8& 57.9 & 61.9 & 1.1\\
  \midrule
& \multicolumn{8}{c}{Backbone (ResNet-101)}       \\
 \midrule
FWB \cite{nguyen2019feature} & ICCV'19 &51.3 & 64.5 & 56.7 & 52.2 & 56.2 & 54.9 & 67.4 & \bf 62.2 & 55.3 & 59.9 & 3.7  \\
DAN \cite{wangfew} & ECCV'20 &54.7 &  68.6 & \bf 57.8 & 51.6 & 58.2 & 57.9 &  69.0 & 60.1 & 54.9 & 60.5 & 2.3  \\
PFNet \cite{tian2020prior} & TPAMI'20 & \bf 60.5 & \bf 69.4 & 54.4 & \bf 55.9 & \bf 60.1 & \bf 62.8 & \bf 70.4 & 54.9 & \bf 57.6 & \bf 61.5 & 1.4 \\
\bf Proposed  & - & 57.0 & 67.2 & 56.1 & 54.3 & 58.7 & 57.3& 68.5 & 61.5 &   56.3 & 60.9 & 2.2\\

\hline
\multicolumn{8}{l}{\scriptsize{* We report the results where no additional unlabeled data is employed.}}\\
\end{tabular}
\label{table:pascal5}
\end{center}
\vspace{-6mm}
\end{table*}

We now report in Table \ref{table:pascal5} an extensive evaluation of all previous works on PASCAL-5$^i$, the most common benchmark in few-shot semantic segmentation.
To make a fair comparison under different feature extractor backbones, we split the table into three groups. The \textit{top} group shows the approaches that rely on VGG-16 as backbone architecture, whereas the methods in the \textit{middle} and \textit{bottom} groups resort to ResNet-50 and ResNet-101 to extract features, respectively. From the reported values, we can observe that the proposed approach outperforms most previous methods, under the same backbone and in both 1- and 5-shot scenarios. Specifically, compared to the second best performing approach based on VGG-16 (i.e., \cite{liu2020crnet}), our method achieves nearly 3\% and 2\% of improvement in 1- and 5-shot, respectively. Furthermore, our approach achieves the best and second best performance across all the methods in the 1- and 5-shot scenarios, respectively, regardless of the backbone architecture. 
These quantitative results demonstrate the strong learning and generalization capabilities of the proposed model in both 1- and 5-shot settings. 

\vspace{-1em}
\subsubsection{Qualitative results.}
\begin{figure*}[h!]
\centering
\includegraphics[width=1\textwidth, trim=0cm 0cm 0cm 0cm]{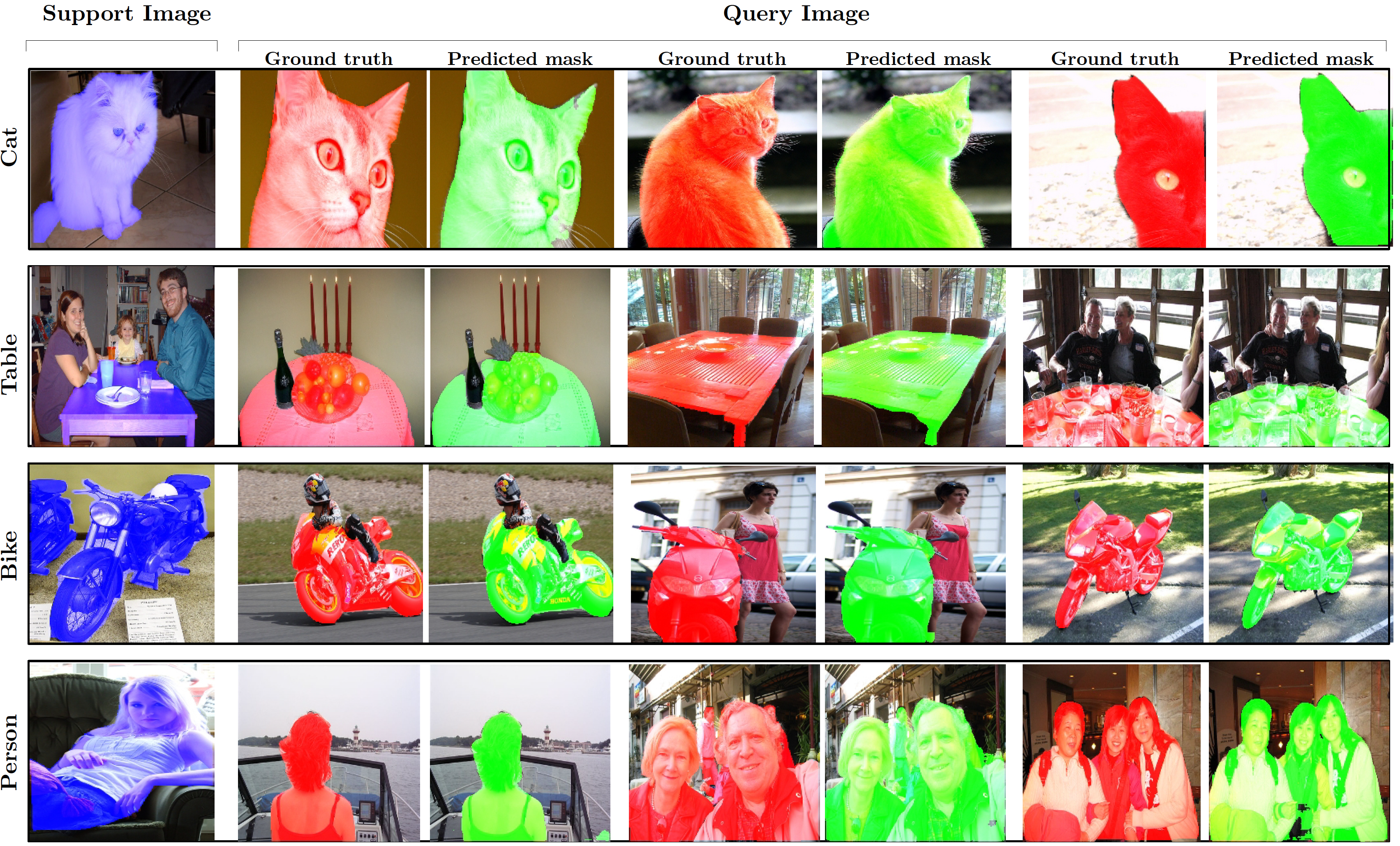}
\caption{Visual results on Pascal-$5^{i}$ in 1-way 1-shot setting using the proposed method. The support set, as well as predictions on several query images with corresponding ground truths are shown.}
\label{fig:pascal_vis}
\vspace{-3mm}
\end{figure*}

We depict visual results of the proposed method on Pascal5$^i$ in Fig \ref{fig:pascal_vis}. Particularly, the support image-mask pair and the segmentation generated by our method for multiple query images, as well as their corresponding ground truths for several categories are shown. 
Without any post-processing step, the proposed model provides satisfying segmentation results on unseen classes with only one annotated support image. It is noteworthy to highlight that the same support image can be employed to segment multiple query images presenting high appearance variability. For example, our model can successfully segment cats (first row of Fig. \ref{fig:pascal_vis}) when only fractions of the target are shown, such as the head (first column) or even a partial head (third column). Looking at other categories, e.g., bike or table, 
we observe that the proposed method can also handle objects viewed from a different perspective or presenting different shapes. This illustrates that our model has a strong ability to successfully generalize to unseen classes from only a handful of labeled examples.

\vspace{-1mm}

\subsubsection{Impact of the multiple scale-space representation fusion strategy}
Logical operations, such as OR, have been typically employed to fuse features from different support images in k-shot segmentation. Even though these operations are straightforward, their result is hard to interpret and they fail to efficiently model the relation between sequential data. Thus, in addition to the results in Table \ref{table:1000Class}, we show in Table \ref{table:1000Class_Sup} the impact of fusing the multiple scale-space representations by both the simple average operation or an additional convolutional layer. Particularly, employing a convolutional layer to combine multiple scale-space representations brings nearly 1\% of improvement compared to the simple average. On the other hand, if we integrate the proposed strategy, the performance is improved by 3\% and 2\%, respectively.  These results demonstrate that our fusion achieves better few-shot segmentation performance.

\begin{table}[h!]
\footnotesize
\begin{center}
\caption{Results of 1-way 1-shot segmentation on the FSS-1000 dataset with different fusion strategies to combine multiple scale-space representations. Best results in bold.}

\label{table:headings}
\begin{tabular}{lccc}
\hline\noalign{\smallskip}
 Method & mIoU\\
\hline
& \multicolumn{1}{c}{1-shot}  \\
\hline
\noalign{\smallskip}
Average  & 77.6 \\
CNN layer & 78.7 \\
\bf Proposed (BConvLSTM)  & \bf 80.8 \\
\hline

\end{tabular}
\label{table:1000Class_Sup}
\end{center}
\vspace{-4.75mm}
\end{table}

\subsubsection{Weakly supervised performance.}
\label{sssec:weak_perf}
We further evaluate the proposed model with weaker forms of annotations, e.g., bounding boxes. As reported in Table \ref{tab:weakSup}, our method achieves comparable performance to full supervision when bounding boxes are available in the support set of novel categories. Furthermore, compared to the very recent PANet architecture \cite{wang2019panet} our model brings 10\% of performance gain in the context of weak supervision. This suggests that our model is able to deal efficiently with noise introduced by bounding box annotations, which ultimately results in more representative class prototypes that approach those obtained by pixel-level annotations.

\begin{table}[h!]
\footnotesize
\centering
\caption{Full supervision vs weak-supervision performance in the 1-shot scenario. Type of supervision in brackets.}
\begin{tabular}{lc|c}
\toprule
\bf  Method   & \multicolumn{2}{c}{\textbf{mIoU}} \\
         & FSS-1000     & PASCAL    \\
         \midrule
Proposed (Pixels)  &  80.8 & 58.0\\
\midrule
Proposed (Bounding boxes)  & 78.2  & 56.4\\
PANet \cite{wang2019panet} (Bounding boxes)  & -  & 45.1\\

\bottomrule
\end{tabular}
    \label{tab:weakSup}
    
\end{table}

\setlength{\tabcolsep}{1.4pt}







\section{Conclusions}

We have presented a novel segmentation network that tackles the challenging problem of few-shot learning from the perspective of reducing the inductive texture bias on CNNs. This contrasts with most prior literature, which focuses on explicitly enhancing the prototypes representation. Particularly, the proposed model presents two novel contributions. First, we integrated a pyramid of Difference of Gaussians to attenuate high-frequency local components in the feature space. Second, to merge information at multiple scale-space representations we reformulated the problem as a sequential task and resorted to bi-directional convolutional LSTMs. 
For evaluation purposes, we have compared the proposed method to prior work, and performed ablations on important elements of our model on public few-shot segmentation benchmarks. Results demonstrated that the proposed model outperforms most prior methods while maintaining a light architecture, achieving a new state-of-the-art performance on several few-shot semantic  segmentation settings.

\bibliographystyle{splncs}
\bibliography{egbib}

\clearpage


\section*{Supplementary Material}

In this supplementary material, we first present extensive details on the datasets used in our experiments. Then, we show additional ablation studies and results that support the satisfactory performance of the proposed method.

\setcounter{section}{0}
\setcounter{table}{0}

\section{Extended details on the employed datasets}
\paragraph{\textbf{PASCAL-5$^i$}.} PASCAL-5$^i$ \cite{shaban2017one} is the most popular few-shot segmentation benchmark, which inherits from the well-known PASCAL dataset \cite{everingham2010pascal}. The images in PASCAL-5$^i$ are split into 4 folds, each having 5 classes, with 3 folds used for training and 1 for evaluation. 
Following the standard procedure in  \cite{shaban2017one,nguyen2019feature}, we employ 1000 support-query pairs randomly sampled in the test split for each class at test time. More details on PASCAL-5$^i$ are provided in \cite{shaban2017one}. 


\paragraph{\textbf{FSS-1000}.} A limitation of PASCAL-5$^i$ is that it contains relatively few distinct tasks, i.e., 20 excluding background and unknown categories. FSS-1000 dataset \cite{wei2019fss} alleviates this issue by introducing a more realistic dataset for few-shot semantic segmentation, which emphasizes the number of object classes rather than the number of images. Indeed, FSS-1000 contains a total of 1000 classes, where only 10 images and their corresponding ground truth for each category are provided. Out of the 1000 classes, 240 are dedicated to the test task and the remaining for training. The FSS-1000 dataset \cite{wei2019fss} only provides pixel-level annotations. Thus, to investigate the effect of using weak annotations in this dataset we generated bounding box annotations. Each bounding box is obtained from one randomly chosen instance mask in each support image. The generated bounding box annotations are provided with the code employed in the experiments.  

\paragraph{\textbf{COCO}} is a challenging large-scale dataset, which contains 80 object categories. Following [26], we choose 40 classes for training, 20 classes for validation and 20 classes for test. 

\section{Importance of the pyramidal setting}

The integration of DoG in our model is strongly inspired by the seminal work in \cite{lowe2004distinctive}. Thus, we followed the recommended setting, which suggests that 5 scale levels gives optimal results. To understand why employing a single DoG with a larger difference of $\sigma$ between the gaussian kernels will not perform at the same level than a pyramid of progressive DoG we need to consider how we recognize images at different distances. When we try to recognize objects that are far away, we might be able to just identify rough details, while fine-grained object details become more clear as the image gets closer. Thus, the level of the scale-space is a key factor when trying to recognize discriminative features in an image. The problem, however, is that the optimal scale-space level to discriminate important features for each object is unknown. By blurring the image with different $\sigma$ values each image represents a different scale-space level, each of them specializing on features at a given 'distance'. In contrast, if we assume a single DoG with a larger difference between the Gaussian kernel variance, intermediate scale-scape levels will be missed. To demonstrate this empirically, we investigated the setting where a single DoG with $\sigma_0$ and $\sigma_4$ is integrated into the CNN. Results reported in Table \ref{tab:sigma_pyr} shows that a single DoG obtains a mIoU value of 77.67 on the FSS-1000 dataset, underperforming by 3\% the pyramidal setting. 

\begin{table}[h!]
    \centering
    \caption{Effect of employing a single DoG with $\sigma_0$ and $\sigma_4$ vs. a pyramidal DoG with progressive $\sigma$ values. 
    Results for 1-shot on the FSS-1000 class dataset.}
    \begin{tabular}{p{0.2\textwidth} p{0.12\textwidth}}
    \hline
         & mIoU \\
    \hline     
    \hline
    
   Single DoG & 77.7 \\
   Pyramidal DoG & 80.8\\
    \hline
    \end{tabular}
    \label{tab:sigma_pyr}
\end{table}

\section{Ablation study on multi-scale fusion features.}Similarly to \cite{zhang2019canet}, we investigated the effect of employing different levels of features, or a combination of those. Particularly, we investigated the three last blocks of VGG-16. 
In our case, \textit{block5} gives the best performance when a single block is used. If multiple blocks are used instead, we observed that combining the three blocks provides the best performance, even though the contribution of the \textit{block4} is marginal compared to the fused features from \textit{block3} and \textit{block5} (+0.26$\%$). The low performance of shallower layers alone can be explained by the fact that they exploit lower-level cues, which are insufficient to properly find object regions. By integrating these with higher-level features, which correspond to object categories, our model can efficiently identify class-agnostic regions on new images. Furthermore, fusion of features at several levels of abstraction can help to handle larger scale object variations. Thus, the final multi-scale model employed in our experiments corresponds to the architecture combining the three last feature blocks.

\begin{table}[h!]
\scriptsize
    \centering
    \caption{Effect of combining different level feature maps in the encoder network. Best result is highlighted in bold.}
    \begin{tabular}{p{0.12\textwidth} p{0.12\textwidth} p{0.12\textwidth} p{0.12\textwidth}}
    \hline
        Block 3 & Block 4 & Block 5 & mIoU \\
    \hline     
    \hline
    
    \checkmark & & & 76.3 \\
    & \checkmark & & 78.3\\
    & &\checkmark & 79.5\\
    \checkmark & \checkmark& & 78.1 \\
    \checkmark & &\checkmark& 80.6\\
    &\checkmark &\checkmark& 79.5\\
    \checkmark & \checkmark& \checkmark & \bf 80.8\\
    
    \hline
    \end{tabular}
    \label{tab:vgg_combination}
\end{table}

\section{Model complexity.}
The functionality of the proposed method in the demand of computational resources is also investigated in this work. Table \ref{tab:complexity} shows the model complexity of several methods, as well as their segmentation results on Pascal5$^i$ for 1-shot. In this table, we include the models that either report their number of parameters or provide reproducible code. We observe that the proposed method is ranked among the lightest methods, while typically achieving the best segmentation performance. Compared to similar methods, in terms of complexity (e.g., co-FCN \cite{rakelly2018conditional}, RPMM\cite{liu2020part} or SG-One \cite{zhang2018sg}), our model brings between 2 and 17\% gain on improvement. 

\begin{table}[h!]
\centering
\caption{Parameter complexity in different approaches and their performance (mIoU) on 1-shot segmentation on PASCAL-$5^{i}$. Methods are ordered based on number of learnable parameters.}
\begin{tabular}{l | c | c}
\toprule
\bf Method & \bf 1-shot mIoU & \bf \#params(M) \\ [0.2ex]
\midrule
OSLSM \cite{shaban2017one}$\diamond$ & 40.8 & 276.7 \\
Meta-Seg \cite{cao2019meta}$\diamond$ & 48.6 & 268.5 \\
AMP \cite{siam2019amp}$\diamond$ & 43.4 & 34.7 \\
co-FCN \cite{rakelly2018conditional}$\diamond$ & 41.1 & 34.2\\
\bf Proposed$^\diamond$ & 58.0 & 22.7 \\
RPMM \cite{yang2020prototype}$^\dagger$ & 56.3 &  19.6 \\
SG-One \cite{zhang2018sg}$\diamond$ & 46.3 & 19.0 \\
CANet \cite{zhang2019canet}$\dagger$ & 55.4 & 19.0 \\
PGNet \cite{zhang2019pyramid}$\dagger$ & 56.0 & 17.2 \\
\bf Proposed$^\ddagger$ & 58.7 & 16.3 \\
PANet \cite{wang2019panet}$\diamond$ & 48.1 & 14.7 \\
PFNet \cite{tian2020prior}$\ddagger$ & 60.1 & 10.8 \\
\hline
\multicolumn{3}{l}{\scriptsize{*Employed architectures: $\diamond$, VGG, $\dagger$ ResNet50, $\ddagger$ ResNet101}}\\
\end{tabular}
    \label{tab:complexity}
\end{table}

\section{Results on COCO}

Table \ref{table:COCO} reports the results for 1- and 5-shot segmentation on COCO dataset. As the backbone architecture plays an important role on the performance of the whole model, we split the results on methods relying on VGG-16 \textit{(top)} and on ResNet \textit{(top)}. From these results we can see that the proposed method achieves the best performance for 1-shot setting on the VGG-16 group, also outperforming a recent approach with ResNet, i.e., \cite{nguyen2019feature}. Regarding the results on 5-shot, our model obtains similar results, but slightly worst, to those obtained by several approaches with ResNet as backbone. This, together with results on FSS-1000 and Pascal5$^i$, supports our hypothesis that removing the texture bias can be more efficient in scenarios with very limited supervision (e.g., 1-shot), where our method consistently achieves the best results across three different datasets (under the exact same conditions, i.e., same architecture as backbone).

\begin{table*}[h!]
\scriptsize
\begin{center}
\caption{Results of 1-way 1-shot and 5-shot segmentation on COCO-20$^i$ data set employing the mean Intersection Over Union (mIoU) metric. Methods are divided according to the backbone used.}
\label{table:headings}
\begin{tabular}{llcccccccccc}
\hline
\noalign{\smallskip}
& & \multicolumn{4}{c}{1-shot} & &\multicolumn{4}{c}{5-shot} & \\
\midrule
Method & & fold$^1$ & fold$^2$ & fold$^3$ & fold$^4$ & Mean & fold$^1$ & fold$^2$ & fold$^3$ & fold$^4$ & Mean \\
\noalign{\smallskip}
 \midrule
 \multicolumn{12}{c}{Backbone (VGG-16)} \\
 \midrule
PANet\cite{wang2019panet} & ICCV'19 & - & - & - & - &20.9 & - & - & -& - &\bf 29.7\\
Proposed& - & 20.2 & 17.8 & 21.6 & 26.8 & \bf 21.6 & 22.6 & 22.0 &24.2 & 31.7 & 25.1\\
 \midrule
 \multicolumn{12}{c}{Backbone (ResNet)} \\
 \midrule
FWB\cite{nguyen2019feature} $\ddagger$ & ICCV'19 & 18.4 & 16.7 & 19.6 & 25.4 & 20.0 & 20.9 & 19.2 & 21.9 & 28.4 & 22.6 \\
OANet \cite{zhao2020objectness} $\ddagger$ & Arxiv'20 & 29.6 & 22.9 & 20.3 & 17.5 & 22.6 & 36.6 & 27.1 & 25.9 & 21.9 & 27.9 \\


DAN \cite{wangfew} $\ddagger$ & ECCV'20 & - & - & - & - &24.4 &- & - & - & - & 29.6\\
RPMM (Baseline) \cite{yang2020prototype} $\dagger$ & ECCV'20 & 25.1 &30.3 & 24.5 &  24.7 & 26.1 & 26.0 & 32.4 & 26.1 & 27.0 & 27.9\\
RPMM \cite{yang2020prototype} $\dagger$ & ECCV'20 & 29.5 & 36.8 & 29.0 & 27.0 & 30.6 & 33.8 & 42.0 & 33.0 & 33.3 &  35.5\\
PPNet* \cite{liu2020part} $\dagger$ & ECCV'20 & 34.5 & 25.4 & 24.3 & 18.6 & 25.7 & 48.3 & 30.9 & 35.7 & 30.2 & 36.2\\
PFNet \cite{tian2020prior} $\ddagger$ & TPAMI'20 & 36.8& 41.8 &38.7 &36.7 &\bf38.5 &40.4 &46.8 &43.2 &40.5 &\bf42.7 \\
\hline
\multicolumn{4}{l}{\scriptsize{Employed architectures: $\dagger$ ResNet50, $\ddagger$ ResNet101}}\\
\end{tabular}
\label{table:COCO}
\end{center}
\end{table*}

\section{Additional visual results}

We include additional qualitative results to assess the performance of our method. First, in Fig. \ref{fig:fss_vis}, visual results on the FSS-1000 class dataset are shown. Similarly to the qualitative examples shown in the main paper, we can observe how our method satisfactorily handles target objects presenting high variability on shape or perspective. This is evident, for example, in the bat images, where our method is able to capture the whole context of a bat flying, while the support image just contained an image of three bats standing in a branch. Then, we also depict failure cases (Fig. \ref{fig:fss_vis_appendix_bad}), where our method does not achieve satisfactory segmentations, or not as good as expected. Typically, these failures come in the form of incomplete segmentations, with small regions of the object not properly identified. The next figure (Fig. \ref{fig:fss_bbox}) depicts the results when a bounding box is employed as supervisory signal in the support sample (depicted in purple). Despite the fact that the support mask is noisy, the results achieved by our method are close to the ground truth masks. This, in addition to the quantitative results reported in Table \ref{tab:weakSup} (main paper), shows that the proposed method, once trained on a base dataset, is robust to noise on the support masks. Last, in Figure \ref{fig:fss_annotations_bbox}, we depict few samples from the FSS-1000 class dataset, with their corresponding ground truth and the generated bounding box annotation.

\setcounter{page}{1}
\setcounter{figure}{0}

\begin{figure*}[h!]
\centering
\includegraphics[width=1\textwidth, trim=0cm 5cm 0cm 0cm]{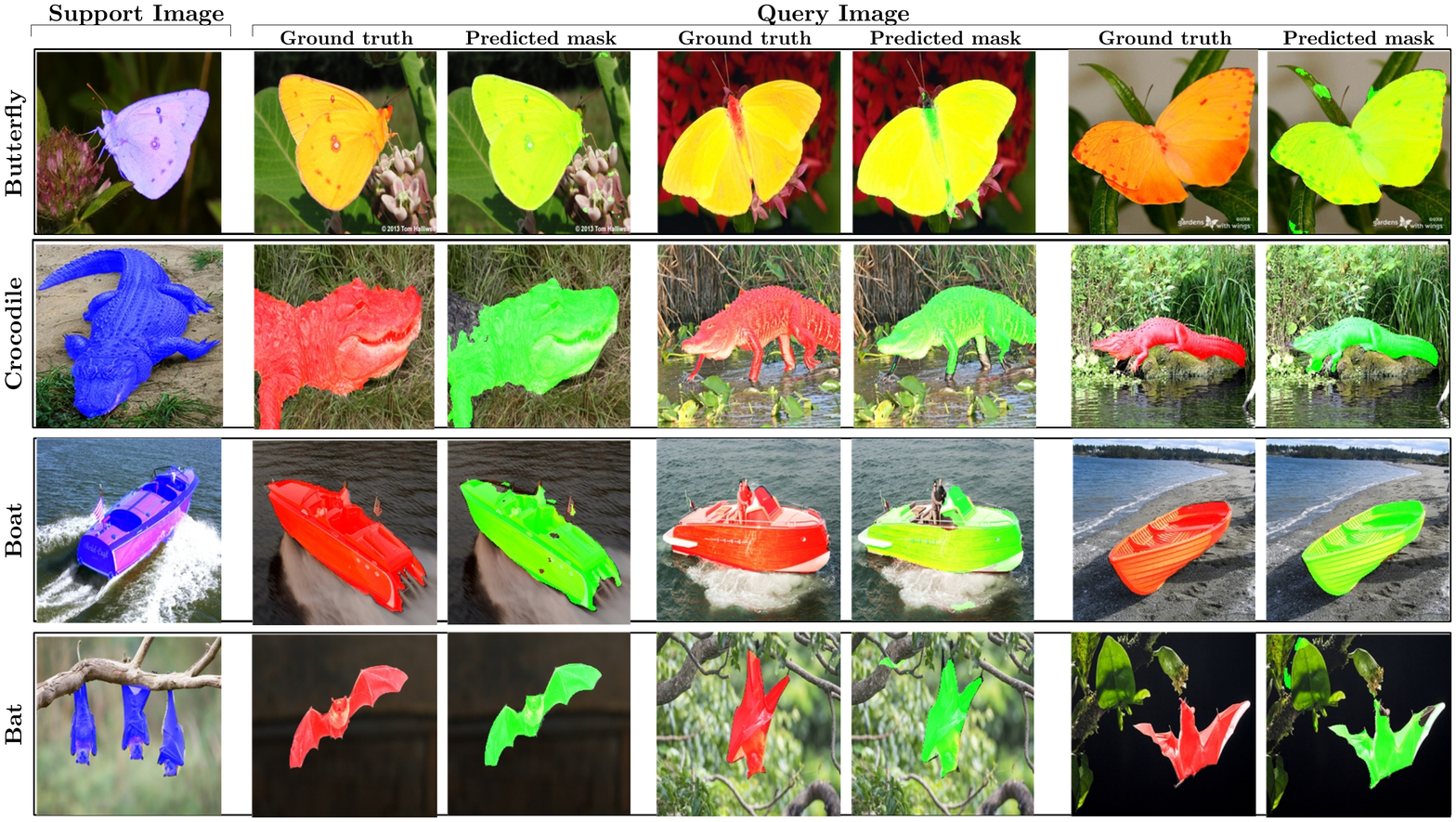}
\caption{Visual results on FSS-1000 class dataset in 1-way 1-shot setting using the proposed method. The support set, as well as predictions on several query images with corresponding ground truths are shown.}
\label{fig:fss_vis}
\end{figure*}

\begin{figure*}[h!]
\centering
\includegraphics[width=1\textwidth]{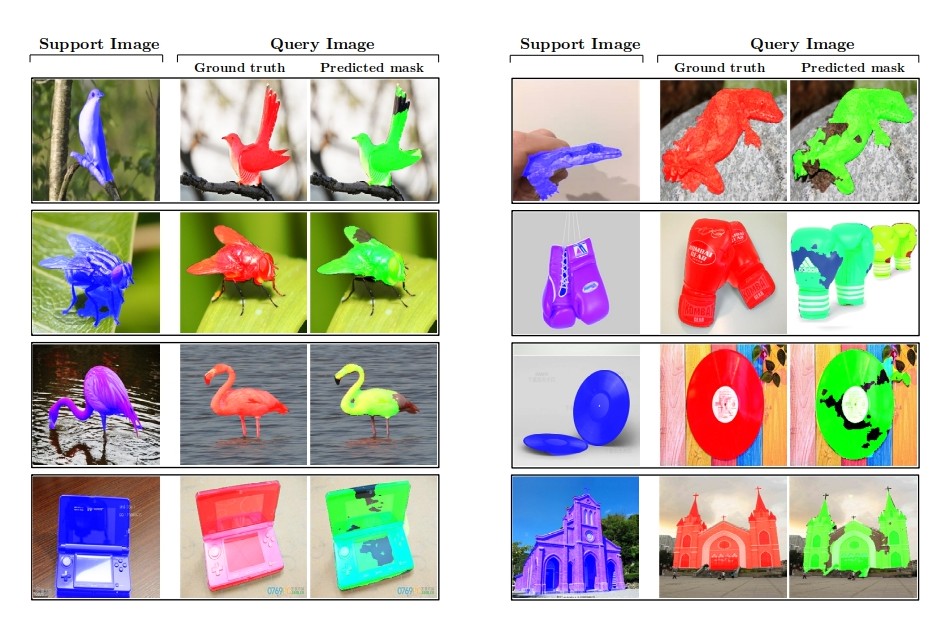}
\caption{Visual examples of \textit{bad} segmentation results on the FSS-1000 class dataset in 1-way 1-shot setting using the proposed method. The support set, as well as predictions on several query images with corresponding ground truths are shown.}
\label{fig:fss_vis_appendix_bad}
\end{figure*}

\begin{figure*}[h!]
\centering
\includegraphics[width=1\textwidth]{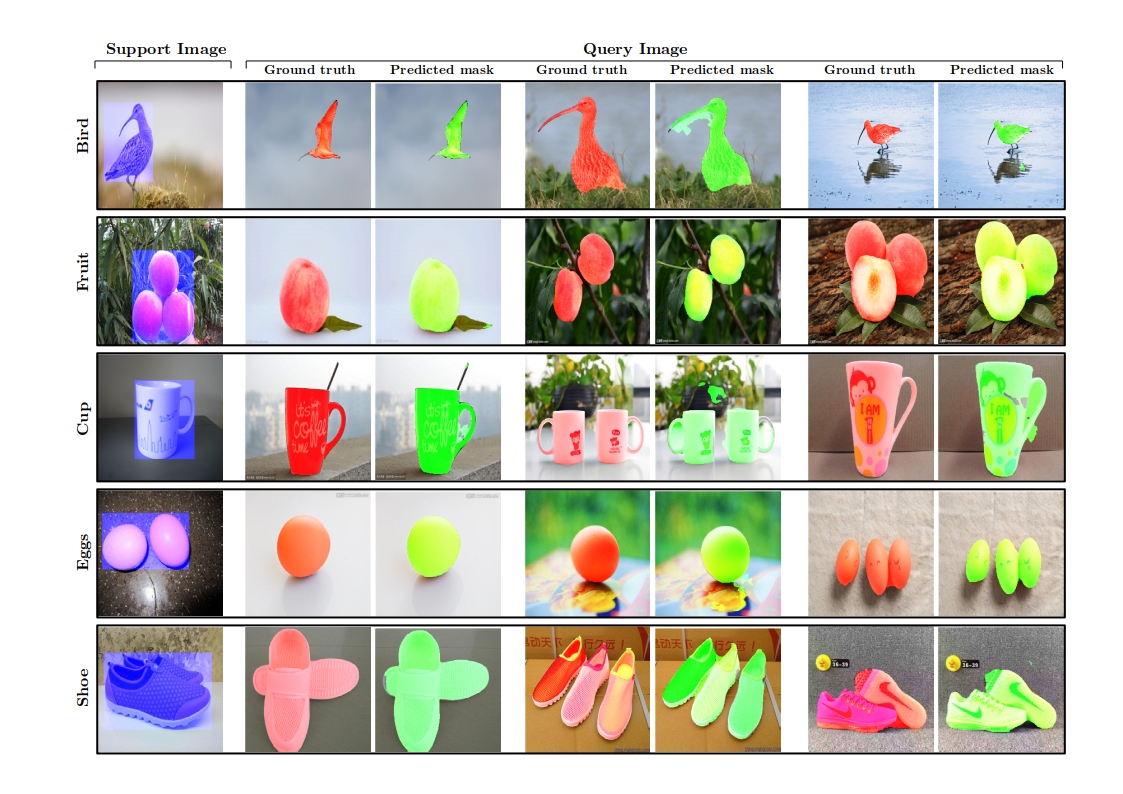}
\caption{Visual examples of segmentation results on the FSS-1000 class dataset in 1-way 1-shot setting using the proposed method \textit{with bounding box annotations}. The support set (i.e., image and its corresponding bounding box annotation), as well as predictions on several query images with corresponding ground truths are shown.}
\label{fig:fss_bbox}
\end{figure*}

\begin{figure*}[h!]
\centering
\includegraphics[width=1\textwidth]{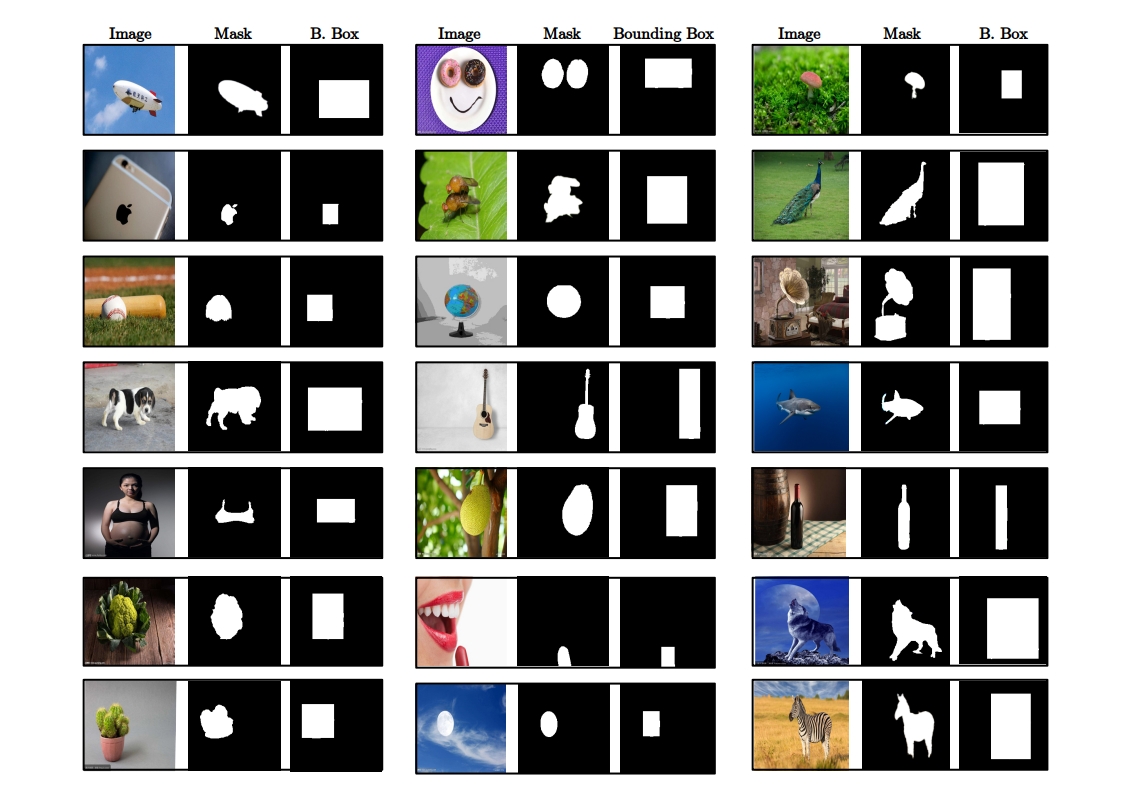}
\caption{Examples of \textit{bounding box annotations} generated on the FSS-1000 class dataset. }
\label{fig:fss_annotations_bbox}
\end{figure*}

\end{document}